\documentclass[runningheads]{llncs}

\usepackage[T1]{fontenc}
\usepackage{slantsc}
\usepackage[utf8]{inputenc}
\usepackage[english]{babel}
\usepackage{microtype}
\usepackage{amsmath}
\usepackage{amssymb}
\usepackage{mathtools}
\usepackage{macros}
\usepackage{hyperref}
\usepackage{booktabs}
\usepackage{multirow}
\usepackage{todonotes}
\usepackage[inline,shortlabels]{enumitem}
\usepackage[caption=false]{subfig}
\usepackage{soul}

\begin{document}
\title{The HyperTrac Project: Recent Progress and Future Research Directions on Hypergraph Decompositions\thanks{This paper is a preprint version of~\cite{DBLP:conf/cpaior/GottlobLLOP20}.}}
\titlerunning{The HyperTrac Project on Hypergraph Decompositions}

\author{
	Georg Gottlob\inst{1}
	\and Matthias Lanzinger\inst{2}
	\and Davide Mario Longo\inst{2}
	\and Cem Okulmus\inst{2}
	\and Reinhard Pichler\inst{2}
}

\authorrunning{G. Gottlob et al.}

\institute{
	University of Oxford, \email{georg.gottlob@cs.ox.ac.uk}
	\and TU Wien, \email{\{mlanzing,dlongo,cokulmus,pichler\}@dbai.tuwien.ac.at}
}

\date{}
\maketitle

\begin{abstract}
Constraint Satisfaction Problems (CSPs) play a central role in many applications in Artificial Intelligence and Operations Research.
In general, solving CSPs is \np-complete. The structure of CSPs is best described by hypergraphs.
Therefore, various forms of hypergraph decompositions have been proposed in the 
literature to identify tractable fragments of CSPs. However, also the computation of a concrete hypergraph decomposition 
is a challenging task in itself. In this paper, we report on recent progress in the study of hypergraph decompositions 
and we outline several directions for future research. 
\end{abstract}

\section{Introduction}
\label{sect:introduction}

Constraint Satisfaction Problems (CSPs) are arguably among the most important problems in Artificial Intelligence with 
a wide range of applications
including diagnosis, planning, natural language processing, machine learning, 
etc.~\cite{BoothTNB16,GangeHS19,RaedtGN10,rossi2006handbook,tsang1993,VerhaegheNPQS19}.
CSPs provide convenient means to formulate combinatorial problems and are, therefore, 
also used in many applications in Operations Research 
spanning scheduling \cite{GeibingerMM19,LaborieRSV18,MSS18},
vehicle routing~\cite{BoothB19,LamHK20,Shaw98},
all kinds of graph problems such as colouring, matching, 
and many other areas~\cite{DBLP:journals/eor/BrailsfordPS99,dechter2003,rossi2006handbook}.

Formally, solving a CSP comes down to model-checking of a first-order formula, 
where the formula only uses the connectives $\exists, \wedge$ but not $\forall,\vee,\neg$.
In this sense, solving CSPs is the equivalent problem to answering Conjunctive Queries (CQs) -- one of the 
most fundamental kinds of queries in the database world, which essentially corresponds to (unnested) SELECT-FROM-WHERE 
queries in the popular database query language SQL or Basic Graph Patterns (BGPs) in the Semantic Web query language SPARQL.
The underlying structure of these problems is best captured by a hypergraph. 
A hypergraph $H =(V,E)$ consists of a set $V$ of vertices and a set $E$ of edges with $E \subseteq 2^V$.
An FO-formula $\phi$ representing a CSP or CQ gives rise to the hypergraph $H =(V,E)$, where $V$ contains the set of variables of $\phi$
and $E$ contains a set $e$ of variables as an edge if and only if there is an atom in $\phi$ whose variables are precisely the ones in $e$.

Solving CSPs and answering CQs  are classical \np-complete problems~\cite{DBLP:conf/stoc/ChandraM77}.
Therefore, there is a long history of research on finding tractable fragments of these problems. 
A natural approach to this task is to search for structural properties of the underlying hypergraph which 
ensure tractability of CSP solving and CQ answering. 
A key result in this area is that CSP instances whose underlying hypergraph is acyclic 
can be solved in polynomial time~\cite{DBLP:conf/vldb/Yannakakis81}. 
Several generalisations of acyclicity have been identified by 
defining various forms of hypergraph {\em decompositions\/},
each associated 
with a specific notion of \emph{width}~\cite{DBLP:journals/ai/GottlobLS00,DBLP:journals/jcss/CohenJG08}.
Intuitively, the width measures how far away a hypergraph is from being acyclic, 
with a width of 1 describing the acyclic hypergraphs. 
The most important forms of decompositions 
are 
{\it hypertree decompositions (HDs)\/}  
\cite{DBLP:journals/jcss/GottlobLS02},  
{\it generalized 
hypertree decompositions (GHDs)}~\cite{DBLP:journals/jcss/GottlobLS02}, and  
{\it fractional hypertree decompositions (FHDs)}~\cite{2014grohemarx}.
These decomposition methods give rise to the following 
notions of width of a hypergraph $H$:
the  {\it hypertree width} $\hw(H)$,  
{\em generalized hypertree width} $\ghw(H)$, 
and {\em fractional hypertree width} 
$\fhw(H)$, 
where $\fhw(H)\leq \ghw(H)\leq \hw(H)$ holds for every 
hypergraph $H$. 
For definitions, see Section~\ref{sect:preliminaries}.

The use of decompositions can significantly speed up  CSP solving and CQ answering. 
In fact, in~\cite{DBLP:conf/sigmod/AbergerTOR16}, a speed-up of up to a factor of 2,500 was reported for the 
CQs studied there. 
Structural decompositions are therefore already being used in commercial products 
and research prototypes, 
both in the CSP area
as well as in database systems~\cite{DBLP:conf/sigmod/AbergerTOR16,DBLP:journals/aicom/AmrounHA16,DBLP:conf/sigmod/ArefCGKOPVW15,DBLP:journals/jetai/HabbasAS15,DBLP:conf/aiia/LalouHA09}.
However, deciding if a given hypergraph $H$ has width $\leq k$ for given $k$ (for one of the width-notions mentioned above) 
is itself a challenging task. Formally, 
for a given width-notion $\width$ and a desired value $k$ of the width, 
we are thus confronted with the following family of problems:

\medskip

\begin{problemdef}[framed]{$\checkp(width, k)$}
  Instance: & A hypergraph $H$.\\
  Question: &  Is $width(H) \leq k?$
\end{problemdef}

\medskip
\noindent
We are also interested in the functional counterpart of these problems where,
in case of a ``yes''-answer, 
a  witnessing decomposition of width $\leq k$ should be output as well.
However, all decision procedures 
recalled in this paper also compute an explicit witness and so
there is little need to distinguish between the decision variant and function variant of this
family of problems.

The $\checkp(\hw, k)$ problem is decidable in polynomial time for any 
fixed~$k$ \cite{DBLP:journals/jcss/GottlobLS02}.  In contrast, 
$\checkp(\ghw, k)$
and $\checkp(\fhw, k)$ are \np-hard already for $k=2$~\cite{2009gottlob,DBLP:conf/pods/FischlGP18}.
Nevertheless, since $\ghw$ and $\fhw$ are in general smaller than $\hw$, 
using GHDs and FHDs
allows, in theory, for even more efficient algorithms for solving CSPs and answering CQs 
than using HDs. This is due to the fact that CSP and CQ algorithms using decompositions have a 
runtime which is exponential in the width. Hence, a smaller width may ultimately pay off even if the 
search for a GHD or FHD is harder than for an HD.
In light of the hardness result, the search
for islands of tractability for $\checkp(\ghw, k)$ and
$\checkp(\fhw, k)$ has, therefore, evolved as an important research goal. 
In total, we see the following three main research directions to further increase
the applicability of decomposition techniques to CSP solving in AI- and OR-applications: 

\begin{itemize}
\item {\em Complexity Analysis.\/} We need to identify restrictions on hypergraphs that guarantee the 
tractability of the 
$\checkp(\ghw, k)$ and $\checkp(\fhw, k)$ problems for fixed $k \geq 1$.
Such restrictions should fulfill two main criteria:
(i) they need to be {\em realistic\/} in the sense that they apply to a large
number of CSPs and/or CQs in real-life applications, and (ii) they
need to be {\em non-trivial\/} in the sense that the restriction itself does not
already imply bounded $\ghw$ or $\fhw$. Trivial restrictions would
be, e.g., bounded treewidth or  acyclicity.

\item {\em Algorithm Design.\/}
The main motivation for identifying tractable fragments is to lay the foundation for algorithms 
which perform well on problem instances that fall into these fragments. 
Consequently, there have been several different approaches to the algorithm design for the \textsc{Check} problem, 
including a top-down construction of the decomposition (as proposed in the original paper on 
HDs~\cite{DBLP:journals/jcss/GottlobLS02}), 
a parallel approach to constructing a decomposition, 
and reductions to other problems such as SMT.
In addition, preprocessing in the form of simplifications of a given hypergraph plays an important role.

\item {\em From Theory to Practice.\/}
To make sure that the decomposition algorithms work well in practice, extensive empirical evaluation is necessary. 
Above all, a good understanding of the hypergraphs occurring in real-world applications is required. Of course, in the real world, 
we do not encounter hypegraphs as such but CSPs and database queries with some underlying hypergraph structure. 
Especially for database queries (to a lesser extent also for CSPs) it has turned out that extracting these hypergraph structures 
is a non-trivial task by itself, since the CQs are somehow ``hidden'' behind the syntax of real-world SQL or SPARQL queries. 
In this paper, we  report on the challenges encountered when setting up a hypergraph benchmark, that has 
already been used for several validation tasks and in competitions. 
\end{itemize}

The paper is organized as follows: in Section 2, we recall some basic notions and results. 
Sections 3 --5 are then devoted to a report on  recent developments in the three main research areas
mentioned above, i.e., ``complexity analysis'', ``algorithm design'', and ``from theory to practice''. 
In Section~6, 
we briefly summarize the current state of affairs and outline promising directions for future research.

\section{Preliminaries}
\label{sect:preliminaries}

We have already introduced 
in Section~\ref{sect:introduction}
hypergraphs as pairs $(V,E)$ consisting of a set $V$ of vertices and 
a set $E$ of edges. It is convenient to assume that $V$ contains no isolated vertices (i.e., vertices not contained in any edge). 
We can then identify a hypergraph $H$ with its edge set $E$ and implicitly assume $V = \bigcup E$. 
A subhypergraph $H' = (V',E')$ of $H$ is then simply obtained by taking a subset $E'$ of $E$
and setting $V' = \bigcup E'$.  The {\em primal graph\/} $G = (W,F)$ of a hypergraph $H = (V,E)$ is 
obtained by setting $W = V$ and defining $F$ such that two vertices form an edge in $G$ if and only if they occur jointly in some 
edge in $E$.

We are interested in the following structural properties of hypergraphs: 
the {\em rank\/} of $H$ is the maximum cardinality of the edges of $H$; the {\em degree\/} of $H$
refers to the maximum number of edges containing a particular vertex.
A class ${\cal C}$ of hypergraphs is said to have {\em bounded rank\/} (or {\em bounded degree\/}) if 
there exists a constant $c$ such that every hypergraph in ${\cal C}$  has rank (or degree) $\leq c$.
In \cite{DBLP:journals/corr/abs-2002-05239}, the notion of $(c,d)$-hypergraphs for integers $c \geq 1$ and $d \geq 0$
was introduced:  $H = (V,E)$ is a $(c,d)$-hypergraph if the intersection of any $c$ distinct edges in $E$ 
has at most $d$ elements, i.e., for every subset $E' \subseteq E$ with $|E'| = c$, we have $|\bigcap E'| \leq d$.  
A class ${\cal C}$ of hypergraphs is said to satisfy the {\em bounded multi-intersection property (BMIP)}
\cite{DBLP:conf/pods/FischlGP18}, if there exist
$c \geq 1$ and $d \geq 0$, such that every $H$ in ${\cal C}$ is a $(c,d)$-hypergraph. As a special
case studied in \cite{DBLP:conf/pods/FischlGLP19,DBLP:conf/pods/FischlGP18}, a 
class ${\cal C}$ of hypergraphs is said to satisfy the {\em bounded intersection property (BIP)},
if there exists $d \geq 0$, such that every $H$ in ${\cal C}$ is a $(2,d)$-hypergraph.

For the definition of hypergraph decompositions and their widths, we need the following notions: 
{\em edge weight functions\/} are of the form $\gamma \colon E \rightarrow [0,1]$. 
We define $B(\gamma)  = \{ v\in V \mid \sum_{e\in E, v\in e} \gamma(e) \geq 1 \}$ 
as the set of all 
vertices {\em covered\/} by $\gamma$ and 
$ \weight(\gamma) = \sum_{e \in E} \gamma(e)$ as the 
{\em weight\/} of $\gamma$. 
The set of edges with non-zero weight 
is called the {\em support\/} of $\gamma$, i.e., 
$\cov(\gamma) = \{e \in E \,\mid\, \gamma(e) > 0\}$. 
We call $\gamma$ 
a \emph{fractional edge cover} of a set $X \subseteq V$
by edges in $E$, 
if $X \subseteq B(\gamma)$.
For $X \subseteq V$, we write 
$\rho_H^*(X)$ to denote the minimum weight over all fractional edge covers of $X$. 
For {\em integral\/} edge covers, the edge weight functions are restricted to  integral values, i.e., 
$\gamma \colon E \rightarrow  \{0,1\}$. 
We write $\rho_H(X)$ to denote the minimum weight over all integral edge covers of~$X$. 
Clearly, $\rho_H^*(X) \leq \rho_H(X)$  holds for any $H = (V,E) $ and $X \subseteq V$. The ratio
$\rho_H(V) / \rho_H^*(V)$ is referred to as the {\em integrality gap\/}.

A tuple $(T, (B_u)_{u \in T})$ is a \emph{tree  decomposition (TD)\/} 
of hypergraph $H =(V,E)$, if $T$ is a tree, 
every $B_u$ is a subset of $V$, and the following conditions are~satisfied:
\begin{enumerate}
\item[(1)] 
 For every edge $e \in E$, there is a node $u$ in $T$,  such that  $e \subseteq B_u$, and
\item[(2)] for every vertex $v \in V$,  $\{u \in T \mid v \in B_u\}$ is connected in $T$.
\end{enumerate}
The vertex sets $B_u$ are usually referred to as the {\em bags\/} of the TD. 
Note that, by slight abuse of notation, we write $u \in T$ to express that $u$ is a node in $T$.

A {\em fractional  hypertree decomposition\/} (FHD) of a hypergraph 
$H=(V,E)$ 
is a tuple 
$\left< T, (B_u)_{u\in T}, (\gamma_u)_{u\in T} \right>$, such that 
$\left< T, (B_u)_{u\in T}\right>$ is a TD of $H$ and 
the 
following condition holds:
\begin{enumerate}
\item[(3)] For each $u\in T$,  $B_u \subseteq  B(\gamma_u)$ holds, i.e., $\gamma_u$ is a fractional 
edge cover of $B_u$.
\end{enumerate}
A {\em generalized  hypertree decomposition\/} (GHD) is an FHD, where $\gamma_u$ is an integral
edge weight function for every $u \in T$. Hence, by condition (3), 
$\gamma_u$ is an integral edge cover of $B_u$. A 
{\em hypertree decomposition\/} (HD) of $H$ is a GHD with the following additional condition (referred to as the 
``special condition'' in \cite{DBLP:journals/jcss/GottlobLS02}):

\begin{enumerate}
\item[(4)] For each $u\in T$, $ V(T_u) \cap B(\gamma_u) \subseteq B_u$, where 
$V(T_u)$ denotes the union of all bags in the subtree of $T$ rooted at $u$.
\end{enumerate}
Because of condition (4), it is important to consider $T$ as a {\em rooted\/} tree
in case of HDs. 
For TDs, FHDs, and GHDs,  the root of $T$ can be arbitrarily chosen or simply ignored.
The {\em width\/} of an FHD, GHD, or HD is defined as 
the maximum weight of the functions $\gamma_u$  over all nodes $u \in T$.
The fractional hypertree width,  generalized hypertree width, and
hypertree width of $H$ (denoted $\fhw(H)$, 
$\ghw(H)$, and $\hw(H)$) is the minimum width over all FHDs, GHDs, and HDs~of~$H$.

We next recall some notions which are of great importance in most of the current decomposition algorithms. 
Consider a hypergraph $H =(V,E)$ and let $S \subseteq V$.  
A set $C$ of vertices with $C \subseteq (V \setminus S)$ is 
\emph{$[S]$-connected} if for any two distinct vertices $v,w \in C$, 
there exists a sequence of vertices
$v = v_0,\dots,v_h = w$ and a sequence of edges
$e_0, \dots, e_{h-1}$ ($h \geq 0$) such that
$\{ v_i, v_{i+1} \} \subseteq ( e_i \setminus S)$, for each
$i \in \{0,\ldots, h-1\}$. 
A set $C \subseteq V$ is an 
\emph{$[S]$-component\/}, if $C$ is maximal $[S]$-connected.
Such a vertex set $S$ that is used to split a hypergraph into components is 
referred to as a \emph{separator}. 
Note that a separator $S$ also gives rise to disjoint subsets of $E$ 
with $E_C := \{e \in E \, \mid \, e \cap C \neq \emptyset\}$. 
The \emph{size of an $[S]$-component} $C$ is defined as the number of edges in $E_C$.
We call $S$ a  \emph{balanced separator} if all $[S]$-components of $H$ have size $\leq \frac{|E|}{2}$.
We say that a TD $(T, (B_u)_{u \in T})$ 
(analogously for FHD, GHD, or HD) is in {\em normal form\/} if 
every internal node $u$ of $T$ satisfies the following condition: 
let $u_1, \dots, u_\ell$ be the child nodes
of $u$. Then, for each $i \in \{1, \dots, \ell\}$, there is a $[B_u]$-component $C_i$ of $H$ with $C_i = V(T_{u_i}) \setminus B_u$, 
where $V(T_{u_i})$ denotes the union of all bags in the subtree of $T$ rooted at $u_i$.

\section{On the Complexity of Checking Widths}
\label{sec:matthias}

The search for tractable fragments of $\checkp(\ghw, k)$ and $\checkp(\fhw ,k)$ has seen significant progress in recent years.
Where there were individual proofs for various properties at first, we now have a overarching theoretical framework and tractability results for highly general properties
that unify the current theory of tractable $\checkp$ fragments. In this section we give a brief overview of this uniform theory and the resulting tractable classes for $\checkp$. The presentation here follows~\cite{DBLP:journals/corr/abs-2002-05239} which is the source of all stated results.

The BMIP will play an important role in our discussion. On the one
hand, the structure of edge intersections has been identified as an
important factor in the complexity of the problem. On the other hand,
it can be argued that real-world problems correspond to
$(c,d)$-hypergraphs with low $c$ and $d$.
An empirical study of these parameters in real-world instances is presented later in Section~\ref{sec:davide}.

\subsection{Decompositions from Candidate Bags}

In hypertree decompositions, the special condition implies a kind of
lower bound on the bags of the decomposition in the sense that certain vertices need to be included in certain bags.
With the generalization
to \ghw and $\fhw$ the special condition is dropped and we lose the
lower bound. This then leaves us with exponentially many possible choices,
even in trivial hypergraphs: e.g., every subset of an edge is a
possible bag.
We will see soon that this exponential number of possible bags is in fact the main challenge
in the construction of polynomial time algorithms for width checking. 

To illustrate this point we consider the computational complexity of
constructing a tree decomposition from a set of \emph{candidate bags}
which is given to the procedure as an input. Through rather straightforward
dynamic programming this is indeed possible in polynomial time.
\begin{theorem}
	\label{thm:frameworkalg}
	Let $H=(V,E)$ be a hypergraph and $\mathbf{S} \subseteq 2^{V}$. There
	exists an algorithm that takes $H$ and $\mathbf{S}$ as an input and
	decides in polynomial time whether there exists a tree decomposition in normal form
	such that for every node $u$ it holds that $B_u \in \mathbf{S}$.
\end{theorem}
Interestingly, the restriction to normal form decompositions is necessary here. Without it the problem is in fact \np-complete~\cite{DBLP:journals/corr/abs-2002-05239}.

We call such a tree decomposition where each bag is from $\mathbf{S}$,
a \emph{candidate tree decomposition} (w.r.t. $\mathbf{S}$). Theorem~\ref{thm:frameworkalg}
thus reduces the check problem to the problem of computing a set of candidate bags $\mathbf{S}$ for hypergraph $H$ such that
there exists a candidate tree decomposition w.r.t. $\mathbf{S}$ if and only if $\ghw(H) \leq k$.
The following example illustrates this idea.

\begin{example}
	\label{ex:1}
	Let $k$ and $r$ be constant integers and let $H=(V,E)$ be a hypergraph with rank at most $r$.
	Let $\mathbf{S} \subseteq 2^{V}$ be the set of all subsets of unions of $k$ edges, i.e.,
	\[
	\mathbf{S} = \bigcup_{E'\in E^k} 2^{\bigcup E'}
	\]
	where $E^k$ contains all $k$ element subsets of $E$.
	Since there are $\binom{|E|}{k}\leq |E|^k$ combinations of $k$ edges
	and each edge has rank at most $r$ we see that $\mathbf{S}$ can be
	computed in time $O(|E|^k2^{kr})$.  Clearly, $\mathbf{S}$ contains
	all bags that can be covered by $k$ edges. It is then easy to see
	that there exists a candidate tree decomposition w.r.t. $\mathbf{S}$
	if and only if $ghw(H)\leq k$.
	
	Thus, computing $\mathbf{S}$ and then using Theorem~\ref{thm:frameworkalg} gives us a polynomial time
	procedure for $\checkp(\ghw, k)$ for hypergraph classes with bounded rank.
	
\end{example}

Note that it is enough to consider only tree decompositions in
Theorem~\ref{thm:frameworkalg}. In our setting it is always possible
to find the respective covers of bags, if they exist, in polynomial
time (recall that $k$ is considered constant). 

\subsection{Computing Candidate Bags}

Example~\ref{ex:1} illustrates the way we can use candidate bags for
tractability results. However, the problem becomes much more complex as soon as we abandon bounded rank since we can no longer
enumerate all (exponentially many) bags. The problem thus shifts to constructing appropriate sets of candidate bags.

This splitting of the problem into constructing candidate tree
decompositions and computing lists of candidate bags then becomes very
convenient. It separates the algorithmic considerations for
constructing a decomposition from the combinatorial problem of
limiting the number of bags. Proving both separately is significantly
simpler than doing both at the same time.

Hence, for a polynomial size list of candidate bags we need to consider a more limited set of decompositions.
In particular, we focus only on \emph{bag-maximal} GHDs in which every bag is made as large as possible.
Bag-maximal GHDs have two important properties: First, there always exists a bag-maximal GHD of $H$ with width $\ghw(H)$. Second, for every edge $e$ and node $u$ in the decomposition we can characterize $e \cap B_u$ (assuming it is not empty) by some covers in the decomposition in the following way
\[
e \cap B_u = \bigcap_{j=1}^\ell e \cap B(\gamma_{u_j})
\]
where $(u_1, \dots, u_\ell)$ is the path from $u$ to the node $u_\ell$ in which $e$ is covered completely, i.e., $e \subseteq B_{u_\ell}$.

However, the length of such paths cannot be bounded in terms of $c$, $d$ and $k$. Instead, we make use of the assumption that we are dealing with a $(c,d)$-hypergraph for constant $c$ and $d$.
Intuitively, we distinguish between two cases based on whether  the intersection along the path intersects $e$ with many ($\geq c$) distinct edges. If so, the intersection is small ($\leq d$) and we can compute all subsets of the intersection. One of them will be $e \cap B_u$. In the other case we can explicitly compute all the
intersections of $e$ with up to $c$ edges, which will again contain $e \cap B_u$. A detailed argument can be found in Section~5 of~\cite{DBLP:journals/corr/abs-2002-05239}.

\begin{theorem}
	\label{thm:ghwtheorem}
	Fix constant $k\geq 1$. For every hypergraph class $\mathcal{C}$ that enjoys the BMIP
	the $\checkp(\ghw, k)$ problem is tractable.
\end{theorem}

We see that $\checkp(\ghw,k)$ is tractable in a wide range of cases. The BMIP properly generalizes many important hypergraph
properties. 
A hypergraph with rank $r$ is a $(1,r)$-hypergraph and a hypergraph
with degree $\delta$ is a $(\delta+1,0)$-hypergraph. Hence, bounded rank and
bounded degree (and bounded intersection) are all simply special cases of
the BMIP.

\subsection{One Step Further: Fractional Hypertree Width}

The restriction to bounded rank is only one part of what made
Example~\ref{ex:1} simple. The other part is that we considered only
$\ghw$. With the step to $\fhw$ it is no longer clear of which sets we want to consider the subsets as more than $k$
edges can be necessary to cover a set of vertices with weight $k$.
In general, the integrality gap for edge cover in hypergraphs is $\Theta(\log |V|)$~\cite{DBLP:journals/dm/Lovasz75}. This means that we would need the union of $k\,\log|V|$ edges to cover every set of vertices $U$ with $\rho^*(U)\leq k$. If we follow the naive approach in Example~\ref{ex:1}, we would thus get a time bound of $O(|E|^{k\log|V|}2^{rk\log|V|})$ which is no longer polynomial.

We see that for $\checkp(\fhw, k)$ we have an additional challenge, bounding the support of fractional edge covers with weight at most $k$. Indeed, in cases where the support is boundable we can, in a sense, reduce the problem to the $\ghw$ case.
This reduction results in Theorem~\ref{thm:rhoq} below.

\begin{definition}
	We say an FHD $\left< T, (B_u)_{u\in T}, (\gamma_u)_{u\in T} \right>$ is \emph{q-limited} if for every node $u$ in $T$ it holds that $|\supp(\gamma_u)|\leq q$.
	Analogous to $\fhw(H)$ we define $\fhw_q(H)$ as the minimum width of all $q$-limited FHDs of $H$.
\end{definition}

\begin{theorem}
	\label{thm:rhoq}
	Fix constants $q$ and $k$. For every hypergraph class $\mathcal{C}$ that enjoys the BMIP
	the $\checkp(\fhw_q, k)$ problem is tractable.
\end{theorem}

Theorem~\ref{thm:rhoq} abstracts away the computation of the tree
decomposition and the computation of the candidate bags. We can show
that \fhw checking is tractable for some class $\mathcal{C}$ if for each
hypergraph $H\in \mathcal{C}$ there exists a constant $q$ such that
$\fhw_q(H) = \fhw(H)$.

\begin{figure}[t]
	\centering
	\subfloat{%
		\includegraphics[height=4em]{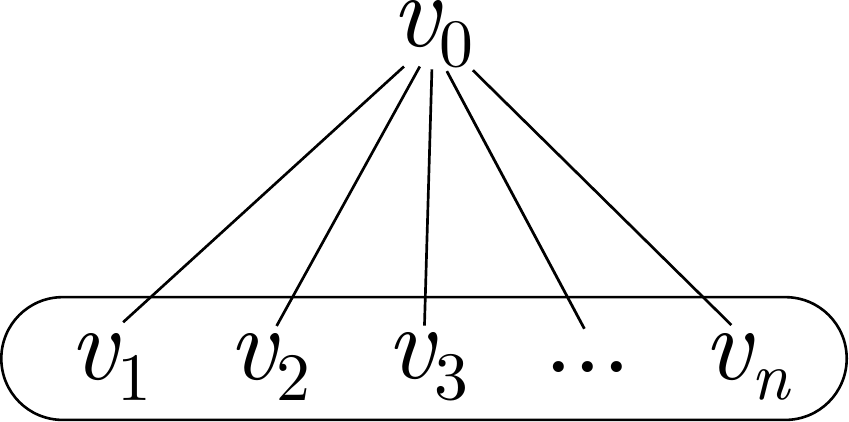}
	}
	\hspace{0.08\textwidth}
	\subfloat{
		$\begin{aligned}[b]
			V(H_n)= &\;\{v_0,v_1,\dots, v_n\} &\\
			E(H_n) = &\;\{ \{v_0,v_i\} \mid 1 \leq i \leq n\}\; \cup &\\
			& \;\{\{v_1,v_2,\dots, v_n\}\}
		\end{aligned}$
	}
	\caption{(2,1)-hypergraphs $H_n$ with large support}
	\label{fig:fracex}
\end{figure}

However, bounding the support of fractional edge covers is a difficult problem.
Consider the $(2,1)$-hypergraphs given in Figure~\ref{fig:fracex}. We have $\rho^*(H_n) = 2- \frac{1}{n}$
where the optimal cover assigns weight $\frac{1}{n}$ to all the edges incident to $v_0$ and
$1-\frac{1}{n}$ to the big edge. That is, the optimal cover has support of size $n$.

The example demonstrates that it is not possible to bound the support of the \emph{optimal} cover, even in $(2,1)$-hypergraphs. However, if we were interested in checking $\fhw \leq 2$ for such a $H_n$, we do not necessarily need to consider the optimal covers but can instead consider slightly heavier covers (still with weight $\leq 2$) for which we can bound the support. In this case it is easy as we can always cover the whole hypergraph with 2 edges that are assigned weight 1.

While the problem is much more complex in general, the main idea stays the same. One can show that for every cover $\gamma$ of weight at most $k$, there exists another weight assignment $\nu$ such that $B(\gamma) \subseteq B(\nu)$, $\weight(\nu) \leq k$, for which the support can be bounded in terms of $k$, $c$, and $d$ if either $c \leq 2$ or $d=0$. In consequence we arrive at the following theorem.

\begin{theorem}
	\label{thm:fhwcheck}
	Fix constants $q$ and $k$. For every hypergraph class $\mathcal{C}$
	that enjoys either bounded degree or bounded intersection the
	$\checkp(\fhw, k)$ problem is tractable.
\end{theorem}

Note that Theorem~\ref{thm:fhwcheck} holds also for classes of bounded rank because
it is a special case of bounded intersection.

\section{Hypergraph Decomposition Algorithms and Systems}
\label{sec:decomposition_algorithms_and_solvers}

In this section we give a  brief overview on recent developments in the field of hypergraph decomposition algorithms. We highlight some selected works which implement decomposition algorithms, i.e., released systems which can produce decompositions and are therefore advancing the practical ability to use decompositions within more complex systems, such as a database management system or a CSP solver. 

\subsection{Hypergraph Preprocessing}

Since simplifications of the input hypergraph are common across many of the presented systems and algorithms, we present here an overview of the techniques so far developed. The common idea is always to reduce the input in such a way that no valid solutions are lost, thus reducing the effective search space. 

Fichte et al.~\cite{DBLP:conf/cp/FichteHLS18} use a number of simplifications, each reducing the size of the SMT encoding: 
\begin{enumerate*}[1)]
	\item the removal of edges contained in other edges, 
	\item splitting $H$ into its biconnected components, and working on each of them separately (for a definition of biconnected components we refer to \cite{DBLP:conf/ecai/GottlobHW02}),
	\item the removal of vertices of degree one and
	\item the removal of simplicial vertices, defined as vertices whose neighbourhood in $H$ forms a clique in the primal graph. This is allowed for the computation of \fhw, since the fractional cover number of this clique then forms a lower bound on the \fhw of H.
\end{enumerate*}
Additional simplifications from the literature are: 
\begin{enumerate*}[1.)]
	\item[5)] the removal of all vertices (bar one) of the same \emph{type}~\cite{DBLP:conf/ijcai/GottlobOP20}, where a \emph{type} of a vertex is the set of all its incident edges and 
	\item[6)]splitting $H$ into its \emph{hinges}~\cite{DBLP:conf/ecai/GottlobHW02}. This is a generalisation of simplification 2). A definition of hinge is found in \cite{DBLP:conf/ecai/GottlobHW02}.
\end{enumerate*}

\subsection{Top-down Construction}

\subsubsection{HD Computation}
We will briefly recall the basic principles of the \detkdecomp algorithm from \cite{DBLP:journals/jea/GottlobS08}, which improves significantly on the first implementation, called \optkdecomp~\cite{DBLP:conf/sebd/LeoneMS02} of the original HD algorithm from \cite{DBLP:journals/jcss/GottlobLS02}.  For a fixed $k \geq 1$, \detkdecomp tries to construct an HD of a hypergraph $H$ in a top-down fashion. Its input is a subhypergraph $H'$ (initially the same as $H$). It produces a new node $u$ (initially serving as the root), then computes the $[\bag{u}]$-components $C_1, \dots, C_\ell$. We define for each component a new hypergraph $H_i = (V_i, E_i)$, where $E_i = \{ e \mid e \cap C_i \not = \emptyset \}$ and $V_i = \bigcup E_i$. Then \detkdecomp recursively searches for an HD of width $\leq k$ for each hypergraph $H_i$. If this succeeds for each $H_i$, then \detkdecomp accepts. If there exists an $H_i$ s.t. no HD of width $\leq k$ can be found, then \detkdecomp backtracks and produces a new node $u$. If all choices for nodes have been exhausted, it rejects.

\subsubsection{Tractable GHD Computation}
\label{ssub:newdetk}
Novel algorithms for solving the aforementioned problem of $\checkp(\ghw, k)$ in polynomial time for $(2,d)$-hypergraphs with low $d$ are presented in \cite{DBLP:conf/pods/FischlGP18}. 
Based on these results, implementations of these algorithms are developed in~\cite{DBLP:conf/pods/FischlGLP19}, built on the basis of \detkdecomp (with the new system aptly named \newdetkdecomp). The source code is publicly available under \url{https://github.com/TUfischl/newdetkdecomp}. 

We proceed to sketch out the tractable \ghw algorithm used in \newdetkdecomp.
As mentioned in Section \ref{sec:matthias}, the main reason for the \np-completeness of the $\checkp(\ghw, k)$ problem is the exponential number of candidate bags. Following the construction from \cite{DBLP:conf/pods/FischlGP18}, \newdetkdecomp explicitly computes intersections of up to $k$ edges, and considers these subsets of edges (called \emph{subedges} in the sequel) as part of the input. From Theorem \ref{thm:ghwtheorem} it follows that this can be done in polynomial time for hypergraph classes with the BIP (as it is a special case of BMIP).
Based on when exactly those subedges are added to the currently considered subhypergraph, two variants were defined, called \globalbip and \localbip. We leave out details here and refer readers to~\cite{DBLP:conf/pods/FischlGLP19}.

\subsection{Parallel Approaches}

\subsubsection{Balanced Separator Algorithm} 
\label{ssub:balsep_ghw_algorithm_}

Yet another \ghw algorithm from~\cite{DBLP:conf/pods/FischlGLP19}, \balsep builds on an observation from~\cite{DBLP:journals/ejc/AdlerGG07}: For any hypergraph $H$ with \ghw  $\leq k$, there exists a balanced separator $S$ with $\rho_H(S) \leq k$. This gives rise to an algorithm which checks for the presence of such separators, and if they cannot be found, can immediately reject. By definition, a balanced separator reduces the size of hypergraphs to be considered by at least half.
This means that \balsep has logarithmically bounded recursion depth, compared with the linear recursion depth of \detkdecomp. This property makes \balsep a promising candidate for a parallel approach to computing \ghw.

\subsubsection{Parallel GHD Computation}
\label{:balancedgo}

On the basis of the Balanced Separator algorithm, a parallel algorithm for computing GHDs is presented in~\cite{DBLP:conf/ijcai/GottlobOP20}, as well as a series of generally applicable algorithmic improvements for computing GHDs. This system, called \balancedgo, is able to decompose nearly twice as many real-world CSP instances within a feasible time, when compared to \newdetkdecomp.
Written in the programming language Go~\cite{donovan2015go}, \balancedgo is available under \url{https://github.com/cem-okulmus/BalancedGo}. 
We proceed to detail this parallel approach below. 

The following generally applicable improvements are presented in~\cite{DBLP:conf/ijcai/GottlobOP20}:
\begin{enumerate*}[1)]
	\item While existing algorithms, such as \detkdecomp make use of heuristics consisting of ordering the edges in such a way as to try out promising separators first, none of the existing methods proved fruitful for speeding up the search for balanced separators. A number of heuristics are considered in~\cite{DBLP:conf/ijcai/GottlobOP20}, ultimately settling on one, which balances out the speed-up of the search against the actual time to compute the heuristic itself, and
	\item the existing implementation of the Balanced Separator algorithm from \cite{DBLP:conf/pods/FischlGLP19} proved to be inefficient w.r.t. considering all relevant subsets of a given separator. Reorganising the way subedges are considered, as well as more effective caching helps to provide significant speed-ups here. 
\end{enumerate*}

The programming language Go has a model of parallelisation inspired by Communicating Sequential Processes of Hoare~\cite{DBLP:journals/cacm/Hoare78}. It is based on light-weight threads, called \emph{goroutines}, which communicate over \emph{channels}. This model minimises, as far as possible, the need for explicit synchronisation. For \balancedgo, there are two areas of parallelism: the search of balanced separators, and the recursive calls. Each is implemented via goroutines, in such a manner as to reduce the need for explicit synchronisation, to enable effective backtracking and utilise existing CPU resources as best as possible. For details we refer to~\cite{DBLP:conf/ijcai/GottlobOP20}.

Finally, a hybrid algorithm is presented~\cite{DBLP:conf/ijcai/GottlobOP20}, which combines the above mentioned parallel Balanced Separator algorithm, with the existing \newdetkdecomp algorithm (extended to compute \ghw as mentioned above): for a constant number $m$ of recursions, it uses the Balanced Separator algorithm. After recursion depth $m+1$ has been reached, it proceeds to use \newdetkdecomp. 
The Balanced Separator algorithm is effective at initially reducing the size of instances, but gets slower as it needs to backtrack more and more often. The \newdetkdecomp algorithm is very effective at quickly computing HDs for smaller instances, or rejecting if no HD of sufficiently low width exists. The hybrid approach therefore combines the best of both worlds.

\subsection{Using Established Solvers and Other Approaches}
\subsubsection{HD and GHD Computation via SMT Encoding}
\label{:frasmt}

A very different approach to compute hypergraph decompositions is utilised by Fichte et al.~\cite{DBLP:conf/cp/FichteHLS18}. Instead of implementing or designing algorithms to compute decompositions, the aim is instead to encode the problem into SMT (SAT modulo Theory) with Linear Arithmetic and then use the SMT solver \emph{Z3}. From this result a provably minimum width FHD can then be constructed. This system, called FraSMT, is available under~\url{https://github.com/daajoe/frasmt}. 

\label{ssub:subsubsection_name}
The basis of the aforementioned encoding is an ordering-based characterisation of \fhw, similar to the well known elimination ordering for treewidth~\cite{DBLP:conf/sofsem/Bodlaender05}. As the elimination ordering has already been used successfully for SAT encodings of treewidth~\cite{DBLP:conf/sat/SamerV09}, it seems natural to investigate a similar approach for \fhw computation. To this end, Fichte et al. define, for a given ordering of the vertices $L = (v_1, \dots, v_n)$, an extension of the hypergraph $H^i_L$, which iteratively constructs and adds a new edge $e_i$, covering all such vertices $v_j$, where $i < j \leq n$ and there exists some edge $e$ in $H^{i-1}_L$ such that $v_i,v_j \in e$.  The \emph{\fhw of H w.r.t. a linear ordering $L$} is then defined as the largest fractional cover number of the vertices $e_i \cup v_i$, for any $v_i \in L$, where only edges in $H$ are considered for the fractional cover. 
Fichte et al. then prove that the $\fhw$ of $H$ is exactly the same as the smallest \fhw of $H$ w.r.t. to any linear ordering.

\label{ssub:smt_encoding_and_symmertry_breaking_}
The above defined ordering is then translated into a formula $F(H,w)$, where $F(H,w)$ is true if and only if $H$ has a linear ordering $L$ such that the \fhw of $H$ is $ \leq w$. For symmetry breaking, Fichte et al. consider the \emph{hyperclique} (defined as cliques in the primal graph), with the highest fractional cover number, and require the vertices of this hyperclique to appear at the end of the ordering, thus reducing the search space.

An extension of FraSMT, is presented by Schidler and Szeider~\cite{DBLP:conf/alenex/SchidlerS20}, called HtdSMT, available under \url{https://github.com/ASchidler/htdsmt}. They extend the above defined encoding to SMT in order to express the special condition of HDs. They won first prize in the PACE 2019 Challenge~\cite{dzulfikar_et_al:LIPIcs:2019:11486}, in the track for the exact computation of \hw of up to 100 instances in less than 30 minutes, beating out a version of \newdetkdecomp.

\label{:htdsmt}

\subsubsection{Approaches Based on the Enumeration of Potential Maximal Cliques}
\label{sub:triangulator_by_korhonen_et_al_}

Korhonen et al.~\cite{DBLP:journals/jea/KorhonenBJ19} present a novel approach to compute \ghw based on the Bouchitt{\'{e}-Todinca (BT) algorithm~\cite{DBLP:journals/siamcomp/BouchitteT01}, which enumerates so-called \emph{potential maximal cliques}(PMC). It is available under \url{https://github.com/Laakeri/Triangulator}. 
While PMC-based approaches were previously used to define a number of algorithms for solving treewidth, minimum fill-in width and other measures, Korhonen et al. are the first to present a practical implementation of the BT algorithm which computes \ghw. Based on their evaluation, it compares quite favourably to \detkdecomp, despite computing \ghw instead of \hw.

\section{From Theory to Practice}
\label{sec:davide}

In this section we discuss the problem of evaluating the quality of decomposition algorithms over real-world instances.
Indeed, while the theory of hypergraph decompositions is well understood and implementations show promising results, very little was known about the typical instances that these should decompose.
To this end, we focus here on the challenges encountered during the development of a benchmark that can be used to reduce the gap between theory and practice.

\subsection{The Need for Benchmarks}
Over the years, the performance evaluation of algorithms and systems tackling variants of the $\checkp$ problem has been conducted against provisional datasets.
When \detkdecomp~\cite{DBLP:journals/jea/GottlobS08} appeared, a collection of CSPs was used to show the superiority of \detkdecomp w.r.t. \optkdecomp.
However, since the dataset lacks CQs, no assessment of \detkdecomp can be made for usage in databases.

Later on, Scarcello et al. proposed in~\cite{DBLP:conf/icde/GhionnaGGS07,DBLP:conf/cikm/GhionnaGS11} a system for the evaluation of SQL queries using hypertree decompositions.
Although their study showed promising results, the dataset used to evaluate the system performance consisted only of a limited set of queries coming from the same source.
Thus, the conclusion cannot be generalized to query answering in general.

These examples highlight the need for a comprehensive, easily extensible, public benchmark.
This could not only be used to evaluate concrete CSP-solving and database systems, but also to empirically test theoretical properties.
Indeed, if a large part of the instances satisfies a certain property, then it is worth developing specialized algorithms exploiting the property.

\subsection{HyperBench: Challenges and Results}
The desire for collecting hypergraphs and comparing algorithm performances led to the implementation of \emph{HyperBench}~\cite{DBLP:conf/pods/FischlGLP19}, which is a comprehensive collection of circa 3000 hypergraphs representing CSPs and CQs.
The hypergraphs and the experimental results are available at \url{http://hyperbench.dbai.tuwien.ac.at}.

\subsubsection{Collecting Instances}
HyperBench altogether contains 3070 hypergraphs divided into three classes: Application-CSPs, Application-CQs, Random.
Table~\ref{tab:hb-instances} shows, for each class, the number of instances and the number of cyclic hypergraphs, i.e., the ones having $\hw \geq 2$.
Out of the 1172 Application-CSPs, 1090 come from XCSP~\cite{xcsp} and 82 were used in previous analyses~\cite{Berg2017maxsat,DBLP:journals/jea/GottlobS08}.
The 535 Application-CQs have been fetched from a variety of sources.
In particular, all the 70 SPARQL queries having $\hw \geq 2$ from~\cite{DBLP:journals/pvldb/BonifatiMT17} and all the cyclic SQL queries from SQLShare~\cite{DBLP:conf/sigmod/JainMHHL16} have been included.
The remaining CQs come from different benchmarks such as the Join Order Benchmark (JOB)~\cite{DBLP:journals/vldb/LeisRGMBKN18} and TPC-H~\cite{tpch}.
The Random class contains 863 random CSP instances from~\cite{xcsp} and 500 random conjunctive queries generated with the tool used for~\cite{DBLP:journals/vldb/PottingerH01}.
This class has the purpose of comparing real application instances to synthetic ones.

\begin{table}
	\caption{Overview of the classes of instances contained in HyperBench~\cite{DBLP:conf/pods/FischlGLP19}.}
	\label{tab:hb-instances}
	\centering
	\setlength{\tabcolsep}{10pt}
	\begin{tabular}{lcc}
		\toprule
		Class				& Num. instances	& \textbf{$\hw \geq 2$} \\
		\midrule
		Application-CSPs	& 1172				& 1172 	\\
		Application-CQs		& 535				& 81 	\\
		Random				& 1363				& 1327 	\\
		\midrule
		Total				& 3070				& 2580	\\	
		\bottomrule
	\end{tabular}
\end{table}

\subsubsection{Obtaining the Hypergraphs}
Collecting CSPs and CQs is only a preliminary step in building a benchmark.
The very next task is the translation of instances into a uniform hypergraph format.
The details of this phase depend on the language in which the instances are written, thus, here, we briefly go over the translation of two specific sets of instances: the CSPs from~\cite{xcsp} and the SQL queries from~\cite{DBLP:conf/sigmod/JainMHHL16}.
While the translation of the first set did not pose any particular challenge, the second one turned out to be rather difficult.

The CSPs fetched from XCSP are encoded in well-structured XML files in which variables and constraints are defined explicitly.
Moreover, an extensive library for parsing the instances, in which most of the process is automatized, is available.
In this case, it is sufficient to redefine the behaviour of some callback methods so that, whenever the program reads a variable, it adds a vertex to the hypergraph, and, whenever it reads a constraint, it adds an edge containing the vertices corresponding to the variables affected by the constraint.

Given its multifaceted nature, the SQLShare dataset poses numerous challenges for the translation of the queries into hypergraphs.
Since it is a collection of databases and handwritten queries by different authors, the original format is highly irregular and requires several refinement phases.
Some of them follow:
\begin{enumerate}
	\item Cleaning the queries from trivial errors that impede parsing.
	\item Extracting table definitions from the databases.
	\item Inferring the definition of undefined tables from the queries.
	\item Resolving ambiguities in the queries semantics, e.g., choosing one definition for the tables that appear with the same name in different databases.
	\item Extracting conjunctive query cores from a complex SQL query, i.e., given a single query, producing a collection of simpler conjunctive queries that can be used to compute the result of the original query.
\end{enumerate}

Particular attention has been devoted to views. 
Indeed, a query that uses views must be expanded first and only afterwards should be translated.
In this way, the resulting hypergraph will accurately reflect the query structure.

\subsubsection{Experiments}
We report on some experiments carried out using HyperBench.
The results reveal a comprehensive picture of the hypergraph characteristics of the collection.
We present aggregate results for the classes in Table~\ref{tab:hb-instances}.

For sake of uniformity, we adapt the terminology of~\cite{DBLP:conf/pods/FischlGLP19} to the one of Section~\ref{sec:matthias}.
A hypergraph $H$ has degree bounded by $\delta$ if and only if $H$ is a $(\delta+1,0)$-hypergraph.
We say $H$ has $c$-multi-intersection size $d$ if $H$ is a $(c,d)$-hypergraph.
In the special case of $c=2$, we talk of intersection size of $H$.
If we do not have in mind any particular $c$, we simply speak of multi-intersection size of $H$.

\begin{table}
	\caption{Percentage of $(c,d)$-hypergraphs with degree $\leq 5$ and $c$-multi-intersection size $\leq 5$, for $c \in \{2,3,4\}$. (6,0)-hypergraphs are the ones with degree at most 5~\cite{DBLP:conf/pods/FischlGLP19}.}
	\label{tab:exp-bip-bmip}
	\centering
	\setlength{\tabcolsep}{10pt}
	\begin{tabular}{lcccc}
		\toprule
		Class				& $(6,0)$-hgs	& $(2,5)$-hgs	& $(3,5)$-hgs	& $(4,5)$-hgs	\\
		&(\%)			&(\%)			&(\%)			&(\%)			\\
		\midrule
		Application-CSPs	& 53.67			& 99.91			& 100			& 100			\\
		Application-CQs		& 81.68			& 100			& 100			& 100			\\
		Random				& 10.12			& 76.82			& 90.17			& 93.62			\\
		\midrule
		Total				& 39.22			& 89.67			& 95.64			& 97.17			\\
		\bottomrule
	\end{tabular}
\end{table}

One of the goals of~\cite{DBLP:conf/pods/FischlGLP19} was to find out whether low (multi-)intersection size is a realistic and non-trivial property.
For the purposes of the study, the value $d=5$ has been identified as a threshold separating low values from high values.
Table~\ref{tab:exp-bip-bmip} shows the percentage of instances having low degree and low $c$-multi-intersection size, for $c \in \{2,3,4\}$.
It can be seen that for each class the amount of instances with low (multi-)intersection is greater than the ones having low degree.
Also, the (multi-)intersection size tends to be (very) small for both CSPs and CQs taken from applications, while it is still reasonably small for random instances.
An additional correlation study between the hypergraph properties establishes that there is no correlation between (multi-)intersection size and $\hw$, thus low (multi-)intersection size does not imply low $\hw$.

After the analysis of structural properties, lower and upper bounds for $\hw$ have been computed for the whole dataset.
For these experiments a timeout of 1 hour was set.
The results are summarized in Table~\ref{tab:hw-leq-5}.
It has been determined that 694 of all 1172 Application-CSPs (59.22\%) have $\hw \leq 5$ and, surprisingly, $\hw \leq 3$ for all Application-CQs.
In total, considering also random instances, 1849 ($60.23\%$) out of 3070 instances have $\hw \leq 5$.
For 1778 of them, the bound on $\hw$ is tight, while for the others the actual value of $\hw$ could be even less.
To conclude, for the vast majority of CSPs and CQs (in particular those from applications), $\hw$ is small enough to allow for efficient CSP solving or CQ answering, respectively.

\begin{table}
	\caption{Number and percentage of instances having $\hw \leq 5$~\cite{DBLP:conf/pods/FischlGLP19}.}
	\label{tab:hw-leq-5}
	\centering
	\setlength{\tabcolsep}{10pt}
	\begin{tabular}{lccc}
		\toprule
		Class				& $\hw \leq 5$ 	& Total		& \% 		\\
		\midrule
		Application-CSPs	& 694			& 1172		& 59.22		\\
		Application-CQs		& 535			& 535		& 100		\\
		Random				& 620			& 1363		& 45.49		\\
		\midrule
		Total				& 1849			& 3070		& 60.23		\\
		\bottomrule
	\end{tabular}
\end{table}

As computing $\ghw$ is more expensive, the algorithms ran on the hypergraphs with small width.
Thus, for all the hypergraphs having $\hw \leq k$ with $k \in \{3,4,5,6\}$, the check $\ghw \leq k-1$ has been performed.
If the algorithm did not timeout and gave either a \emph{yes} or \emph{no} answer, we say the instance is \emph{solved}.
Though it is known that, for each hypergraph $H,\, \hw(H) \leq 3 \cdot \ghw(H) + 1 $ holds~\cite{DBLP:journals/ejc/AdlerGG07}, surprisingly it turns out that 98\% of the solved instances, which form 57\% of all instances, have identical values of $\hw$ and $\ghw$.

\subsection{Further Uses of HyperBench}
Since its publication, HyperBench has been used in several ways in the world of decomposition techniques.
As already discussed, the restrictions defining tractable fragments of variants of the $\checkp$ problem presented in Section~\ref{sec:matthias} have been already investigated in~\cite{DBLP:conf/pods/FischlGLP19}.
Moreover, it has been used to gain an understanding of the differences between $\hw$ and $\ghw$ in real-world CSPs and CQs. 

In~\cite{korhonen2019potential}, the edge clique cover size of a graph is identified as a parameter allowing fixed-parameter-tractable algorithms for enumerating potential maximal cliques.
The latter can be used to compute exact $\ghw$ and $\fhw$.
An edge clique cover of a graph is a set of cliques of the graph that covers all of its edges.
In case of a CSP with $n$ variables and $m$ constraints, the set of constraints is an edge clique cover of the underlying (hyper)graph.
Thus, this property can be exploited for CSPs having $n > m$ and HyperBench has been used to verify that it happens in circa 23\% of the instances.

HyperBench has also been used in the PACE 2019 Challenge~\cite{dzulfikar_et_al:LIPIcs:2019:11486} to compare the performance of several HD solvers.
The challenge dedicated two tracks to HDs:
in the \emph{exact} track, the participants had to compute $\hw(H)$ for as many hypergraphs as possible, while in the \emph{heuristic} track, the task was to compute a decomposition with low $\hw$ in short time.

\section{Conclusion and Future Work}
\label{sect:conclusion}

In this paper, we have reported on  recent progress in the research on hypergraph decompositions. 
This progress has several facets: we now have a fairly good understanding of the complexity of constructing various kinds of 
decompositions. In particular, we have seen several structural restrictions on hypergraphs which make the 
$\checkp(\ghw, k)$ and $\checkp(\fhw, k)$  problems
tractable, such as low rank, low degree, small intersection and, in case of $\ghw$, also small multi-intersection. 
On the algorithmic side, several different approaches have been
proposed for the computation of concrete hypergraph decompositions -- either for some desired upper bound $k$ on the width
or with minimum width. The main categories of these decomposition algorithms are the ``classical'' top-down 
construction of a decomposition (as suggested in \cite{DBLP:journals/jcss/GottlobLS02}), 
a parallel approach, and the reduction to other problems such as SMT. Finally, we have recalled the work on the HyperBench 
benchmark, which has already been used for the empirical evaluation of several implementations of decomposition  
algorithms and 
which has also allowed us to get a realistic picture as to which structural properties the hypergraphs underlying CSPs and CQs
in practice typically  have.

There are several promising directions of future research in this area. 
As mentioned in 
Section~\ref{sect:introduction}, decomposition techniques have already been introduced into
research prototypes and first commercial products. 
Our aspiration is to see decomposition techniques incorporated  more widely also into mainstream systems 
-- both in the CSP world 
and in the database world. Another kind of system where decomposition techniques may have a lot of potential are 
integer programming solvers. Indeed, integer programs are readily modelled as hypergraphs whose vertices correspond to the 
variables in the integer program and the non-zero entries in each row are represented by an edge.
The application of hypergraph decompositions, in particular, to sparse integer programs seems very promising to us. The topic has been touched on
in \cite{korimort2003} but a deeper investigation is missing to date. 
Another important direction for future work is to study the dynamics of decompositions when the corresponding CSP or CQ is 
slightly modified. In such a case, does the entire decomposition have to be re-computed from scratch or can it 
be obtained from the ``old'' one via suitable modifications?

\section*{Acknowledgements}
This work was supported by the Austrian Science Fund (FWF):P30930-N35 in the context of the project ``HyperTrac''. 
Georg Gottlob is a Royal Society Research Professor and acknowledges support by the Royal Society 
for the present work in the context of the project ``RAISON DATA'' 
(Project reference: RP\textbackslash R1\textbackslash 201074).
Davide Mario Longo's work was also supported by the FWF project W1255-N23.}

\bibliographystyle{splncs04}
\bibliography{cpaior20}

\end{document}